\documentclass[11pt,a4paper]{article}
\usepackage[hyperref]{naaclhlt2018}
\usepackage{times}
\usepackage{latexsym}

\usepackage{url}

\usepackage{graphicx}
\usepackage{amsmath}
\usepackage{amsfonts}
\graphicspath{ {figs/} }

\aclfinalcopy 
\def\aclpaperid{1045} 

\title{Multi-Reward Reinforced Summarization with Saliency and Entailment}

\author{Ramakanth Pasunuru \and Mohit Bansal \\
  UNC Chapel Hill \\
  {\tt \{ram, mbansal\}@cs.unc.edu} \\
 }

\date{}

\begin{document}
\maketitle
\begin{abstract}
Abstractive text summarization is the task of compressing and rewriting a long document into a short summary while maintaining saliency, directed logical entailment, and non-redundancy. In this work, we address these three important aspects of a good summary via a reinforcement learning approach with two novel reward functions: ROUGESal and Entail, on top of a coverage-based baseline. The ROUGESal reward modifies the ROUGE metric by up-weighting the salient phrases/words detected via a keyphrase classifier. The Entail reward gives high (length-normalized) scores to logically-entailed summaries using an entailment classifier. Further, we show superior performance improvement when these rewards are combined with traditional metric (ROUGE) based rewards, via our novel and effective multi-reward approach of optimizing multiple rewards simultaneously in alternate mini-batches. Our method achieves the new state-of-the-art results (including human evaluation) on the CNN/Daily Mail dataset as well as strong improvements in a test-only transfer setup on DUC-2002.

\end{abstract}

\section{Introduction}
\label{sec-intro}

Abstractive summarization, the task of generating a natural short summary of a long document, is more challenging than the extractive paradigm, which only involves selection of important sentences or grammatical sub-sentences~\cite{jing2000sentence,knight2002summarization,clarke2008global,filippova2015sentence}. Advent of sequence-to-sequence deep neural networks and large human summarization datasets~\cite{hermann2015teaching,nallapati2016abstractive} made the abstractive summarization task more feasible and accurate,  with recent ideas ranging from copy-pointer mechanism and redundancy coverage, to metric reward based reinforcement learning~\cite{rush2015neural,chopra2016abstractive,ranzato2015sequence,nallapati2016abstractive,see2017get}.

A good abstractive summary requires several important properties, e.g., it should choose the most salient information from the input document, be logically entailed by it, and avoid redundancy. Coverage-based models address the latter redundancy issue~\cite{Suzuki2016Summ,nallapati2016abstractive,see2017get}, but there is still a lot of scope to teach current state-of-the-art models about  saliency and logical entailment.
Towards this goal, we improve the task of abstractive summarization via a reinforcement learning approach with the introduction of two novel rewards: `ROUGESal' and `Entail', and also demonstrate that these saliency and entailment skills allow for better generalizability and transfer.

Our ROUGESal reward gives higher weight to the important, salient words in the summary, in contrast to the traditional ROUGE metric which gives equal weight to all tokens. These weights are obtained from a novel saliency scorer, which is trained on a reading comprehension dataset's answer spans  to give a saliency-based probability score to every token in the sentence. Our Entail reward gives higher weight to summaries whose sentences logically follow from the ground-truth summary. Further, we also add a length normalization constraint to our Entail reward, to importantly avoid misleadingly high entailment scores to very short sentences.

Empirically, we show that our new rewards with policy gradient approaches perform significantly better than a cross-entropy based state-of-the-art pointer-coverage baseline. We show further performance improvements by combining these rewards via our novel multi-reward optimization approach, where we optimize multiple rewards simultaneously in alternate mini-batches (hence avoiding complex scaling and weighting issues in reward combination), inspired from how humans take multiple concurrent types of rewards (feedback) to learn a task. Overall, our methods achieve the new state-of-the-art (including human evaluation) on the CNN/Daily Mail dataset as well as strong improvements in a test-only transfer setup on DUC-2002. Lastly, we present several analyses of our model's saliency, entailment, and abstractiveness skills.

\section{Related Work}
\label{Related Work}

Earlier summarization work was based on extraction and compression-based approaches~\cite{jing2000sentence,knight2002summarization,clarke2008global,filippova2015sentence}, with more focus on graph-based~\cite{giannakopoulos2009automatic,ganesan2010opinosis} and discourse tree-based~\cite{gerani2014abstractive} models. Recent focus has shifted towards abstractive, rewriting-based summarization based on parse trees~\cite{cheung2014unsupervised,wang2016sentence}, Abstract Meaning Representations~\cite{liu2015toward,dohare2017text}, and neural network models with pointer-copy mechanism and coverage~\cite{rush2015neural,chopra2016abstractive,Chen2016DistractionBasedNN,nallapati2016abstractive,see2017get}, as well as reinforce-based metric rewards~\cite{ranzato2015sequence,paulus2017deep}. 
We also use reinforce-based models, but with novel reward functions and better simultaneous multi-reward optimization methods.

Recognizing Textual Entailment (RTE), the task of classifying two sentences as entailment, contradiction, or neutral, has been used for Q\&A and IE tasks~\cite{harabagiu2006methods,dagan2006pascal,lai2014illinois,jimenez2014unal}. Recent neural network models and large datasets~\cite{bowman2015large,williams2017broad} enabled stronger accuracies. Some previous work~\cite{mehdad2013abstractive,gupta2014text} has explored the use of RTE by modeling graph-based relationships between sentences to select the most non-redundant sentences for summarization. Recently,~\newcite{pasunuru2017reinforced} improved video captioning with entailment-corrected rewards. We instead directly use  multi-sentence entailment knowledge (with additional length constraints) as a separate RL reward to improve abstractive summarization, while avoiding their penalty hyperparameter tuning.

For our saliency prediction model, we make use of the SQuAD reading comprehension dataset~\cite{rajpurkar2016squad}, where the answer spans annotated by humans for important questions, serve as an interesting and effective proxy for keyphrase-style salient information in summarization. Some related previous work has incorporated document topic/subject classification~\cite{isonuma2017extractive} and webpage keyphrase extraction~\cite{zhang2004world} to improve saliency in summarization.
Some recent work~\newcite{subramanian2017neural} has also used answer probabilities in a document to improve question generation.

\section{Models}

\subsection{Baseline Sequence-to-Sequence Model}
Our abstractive text summarization model is a simple sequence-to-sequence single-layer bidirectional encoder and unidirectional decoder LSTM-RNN,
with attention~\cite{bahdanau2014neural}, pointer-copy, and coverage mechanism -- please refer to~\newcite{see2017get} for details.

\subsection{Policy Gradient Reinforce}
\begin{figure}
\centering
\includegraphics[width=0.9\linewidth]{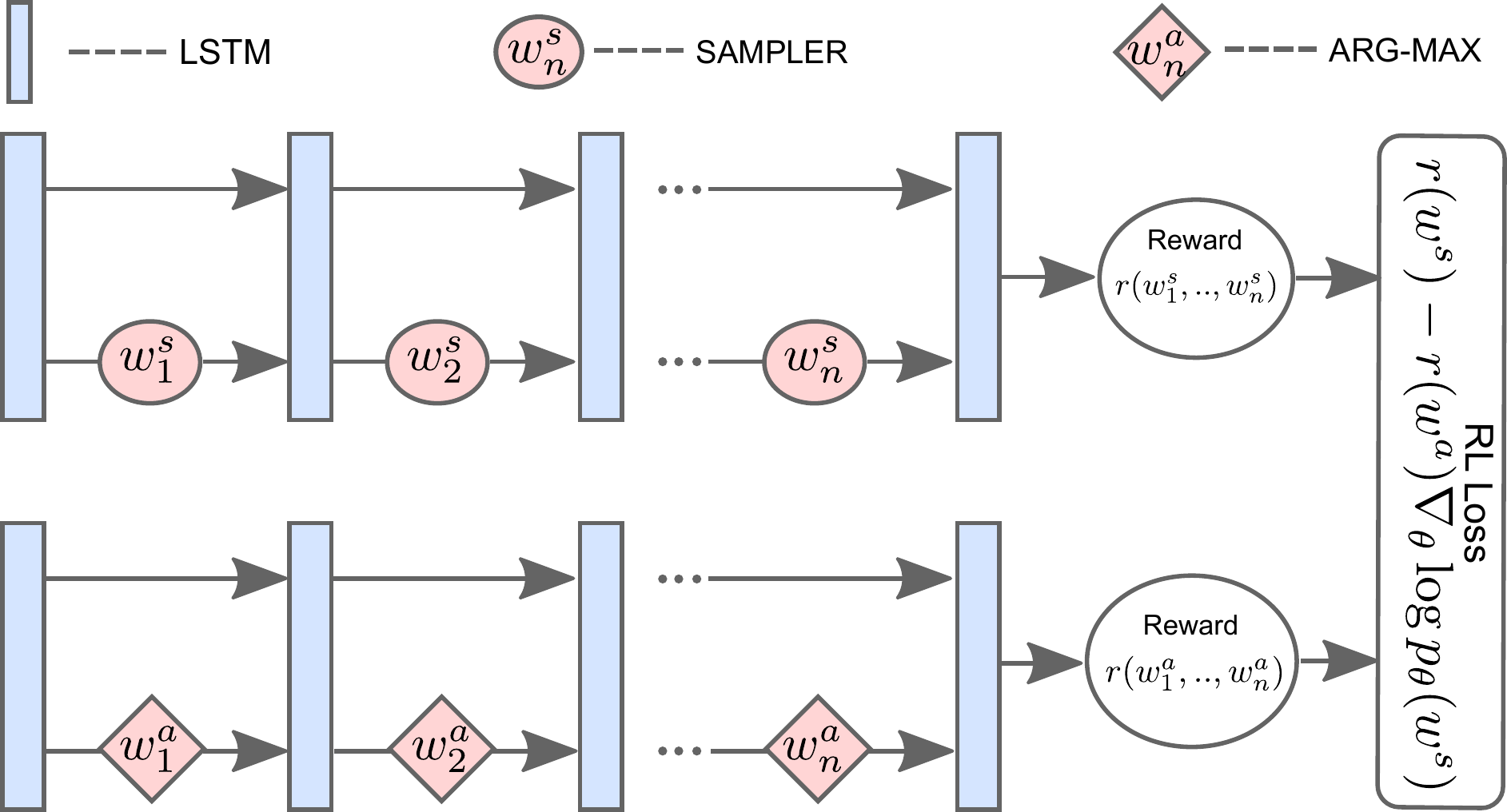}
\vspace{-10pt}
\caption{Our sequence generator with RL training.\vspace{-5pt}
}
\vspace{-15pt}
\label{fig:rl-model}
\end{figure}

Traditional cross-entropy loss optimization for sequence generation has an exposure bias issue and the model is not optimized for the evaluated metrics~\cite{ranzato2015sequence}. Reinforce-based policy gradient approach addresses both of these issues by using its own distribution during training and by directly optimizing the non-differentiable evaluation metrics as rewards. We use the REINFORCE algorithm~\cite{williams1992simple,zaremba2015reinforcement} to learn a policy $p_\theta$ defined by the model parameters $\theta$ to predict the next action (word) and update its internal (LSTM) states. We minimize the loss function $L_{\textrm{RL}} = -\mathbb{E}_{w^s \sim p_\theta} [r(w^s)]$, where $w^s$ is the sequence of sampled words with $w^s_t$ sampled at time step $t$ of the decoder. The derivative of this loss function with approximation using a single sample along with variance reduction with a bias estimator is:
\begin{equation}
\nabla_\theta L_{\textrm{RL}} = -(r(w^s)-b_e) \nabla_\theta \log p_\theta(w^s)
\end{equation}
There are several ways to calculate the baseline estimator; we employ the effective SCST approach~\cite{rennie2016self}, as depicted in Fig.~\ref{fig:rl-model}, where $b_e=r(w^a)$, is based on the reward obtained by the current model using the test time inference algorithm, i.e., choosing the arg-max word $w^a_t$ of the final vocabulary distribution at each time step $t$ of the decoder. We use the joint cross-entropy and reinforce loss so as to optimize the non-differentiable evaluation metric as reward while also maintaining the readability of the generated sentence~\cite{wu2016google,paulus2017deep,pasunuru2017reinforced}, which is defined as $L_{\textrm{Mixed}} = \gamma L_{\textrm{RL}} + (1-\gamma)L_{\textrm{XE}}$, where $\gamma$ is a tunable hyperparameter.
 
\subsection{Multi-Reward Optimization}
\label{subsec:multi-reward-optimization}
Optimizing multiple rewards at the same time is important and desired for many language generation tasks. One approach would be to use a weighted combination of these rewards, but this has the issue of finding the complex scaling and weight balance among these reward combinations. To address this issue, we instead introduce a simple multi-reward optimization approach inspired from multi-task learning, where we have different tasks, and all of them share all the model parameters while having their own optimization function (different reward functions in this case). If $r_1$ and $r_2$ are two reward functions that we want to optimize simultaneously, then we train the two loss functions of Eqn.~\ref{eq:multi-reward-functions} in alternate mini-batches.
\vspace{-10pt}
\begin{equation}
\label{eq:multi-reward-functions}
\begin{aligned}
L_{\textrm{RL}_1} = -(r_1(w^s)-r_1(w^a)) \nabla_\theta \log p_\theta(w^s) \\
L_{\textrm{RL}_2} = -(r_2(w^s)-r_2(w^a)) \nabla_\theta \log p_\theta(w^s) 
\end{aligned}
\end{equation}

\section{Rewards}
\label{sec:rewards}

\paragraph{ROUGE Reward}
The first basic reward is based on the primary summarization metric of ROUGE package~\cite{lin2004rouge}. Similar to~\newcite{paulus2017deep}, we found that ROUGE-L metric as a reward works better compared to ROUGE-1 and ROUGE-2 in terms of improving all the metric scores.\footnote{For the rest of the paper, we mean ROUGE-L whenever we mention ROUGE-reward models.} Since these metrics are based on 
simple phrase matching/n-gram overlap, they do not focus on important summarization factors such as salient phrase inclusion and directed logical entailment. Addressing these issues, we next introduce two new reward functions.

\begin{figure}
\centering
\includegraphics[width=0.90\linewidth]{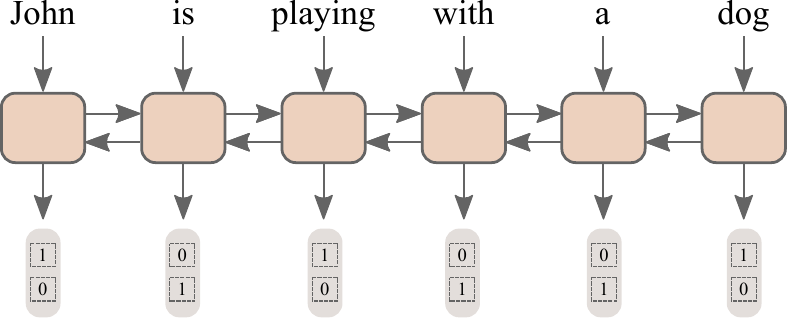}
\vspace{-10pt}
\caption{Overview of our saliency predictor model.}
\vspace{-15pt}
\label{fig:saliency-predictor}
\end{figure}

\paragraph{Saliency Reward}
\label{subsec:saliency-rewards}
ROUGE-based rewards have no knowledge about what information is salient in the summary, and hence we introduce a novel reward function called `ROUGESal' which gives higher weight to the important, salient words/phrases when calculating the ROUGE score (which by default assumes all words are equally weighted). To learn these saliency weights, we train our saliency predictor on sentence and answer spans pairs from the popular SQuAD reading comprehension dataset~\cite{rajpurkar2016squad}) (Wikipedia domain), where we treat the human-annotated answer spans (avg. span length $3.2$) for important questions as representative salient information in the document. As shown in Fig.~\ref{fig:saliency-predictor}, given a sentence as input, the predictor assigns a saliency probability to every token, using a simple bidirectional encoder with a \emph{softmax} layer at every time step of the encoder hidden states to classify the token as salient or not. Finally, we use the probabilities given by this saliency prediction model as weights in the ROUGE matching formulation to achieve the final ROUGESal score (see appendix for details about our ROUGESal weighted precision, recall, and F-1 formulations).

\paragraph{Entailment Reward}
\label{subsec:entailment-rewards}
A good summary should also be logically entailed by the given source document, i.e., contain no contradictory or unrelated information.~\newcite{pasunuru2017reinforced} used entailment-corrected phrase-matching metrics  (CIDEnt) to improve the task of video captioning; we instead directly use the entailment knowledge from an entailment scorer and its multi-sentence, length-normalized extension as our `Entail' reward, to improve the task of abstractive text summarization. We train the entailment classifier~\cite{parikh2016decomposable} on the SNLI~\cite{bowman2015large} and Multi-NLI~\cite{williams2017broad} datasets and calculate the entailment probability score between the ground-truth (GT) summary (as premise) and each sentence of the generated summary (as hypothesis), and use  avg. score as our Entail reward.\footnote{Since the GT summary is correctly entailed by the source document, we directly (by transitivity) use this GT as premise for easier (shorter) encoding. We also tried using the full input document as premise but this didn't perform as well (most likely because the entailment classifiers are not trained on such long premises; and the problem with the sentence-to-sentence avg. scoring approach is discussed below).\\
We also tried summary-to-summary entailment scoring (similar to ROUGE-L) as well as pairwise sentence-to-sentence avg. scoring, but we found that avg. scoring of ground-truth summary (as premise) w.r.t. each generated summary's sentence (as hypothesis) works better (intuitive because each sentence in generated summary might be a compression of multiple sentences of GT summary or source document).} Finally, we add a length normalization constraint to avoid very short sentences achieving misleadingly high entailment scores: 
\vspace{-5pt}
\begin{equation}
\begin{small}
\label{eq:ent-length-norm}
\text{Entail} = \text{Entail} \times \frac{\text{\#tokens in generated summary}}{\text{\#tokens in reference summary}}
\end{small}
\end{equation}

\section{Experimental Setup}
\label{sec-setup}

\subsection{Datasets and Training Details}
\label{subsec:datasets}
CNN/Daily Mail dataset~\cite{hermann2015teaching,nallapati2016abstractive} is a collection of online news articles and their summaries. We use the non-anonymous version of the dataset as described in~\newcite{see2017get}. For test-only generalization experiments, we use the DUC-2002 single document summarization dataset\footnote{\scriptsize{\url{http://www-nlpir.nist.gov/projects/duc/guidelines/2002.html}}}.
For entailment reward classifier, we use a combination of the full Stanford Natural Language Inference (SNLI) corpus~\cite{bowman2015large} and the recent Multi-NLI corpus~\cite{williams2017broad} training datasets.
For our saliency prediction model, we use the Stanford Question Answering (SQuAD) dataset~\cite{rajpurkar2016squad}. All dataset splits and other training details (dimension sizes, learning rates, etc.) for reproducibility are in appendix.

\subsection{Evaluation Metrics}
\label{subsec:evaluation-metrics}
We use the standard ROUGE package~\cite{lin2004rouge} and Meteor package~\cite{banerjee2005meteor} for reporting the results on all of our summarization models. Following previous work~\cite{chopra2016abstractive,nallapati2016abstractive,see2017get}, we use the ROUGE full-length F1 variant.

\noindent\textbf{Human Evaluation Criteria:}
We also performed human evaluation of summary \emph{relevance} and \emph{readability}, via Amazon Mechanical Turk (AMT). We selected human annotators that were located in the US, had an approval rate greater than $98\%$, and had at least $10,000$ approved HITs. For the pairwise model comparisons discussed in Sec.~\ref{sec-results}, we showed the annotators the input article, the ground truth summary, and the two model summaries (randomly shuffled to anonymize model identities) -- we then asked them to choose the better among the two model summaries or choose `Not-Distinguishable' if both summaries are equally good/bad. Instructions for relevance were based on the summary containing salient/important information from the given article, being correct (i.e., avoiding contradictory/unrelated information), and avoiding redundancy. Instructions for readability were based on the summary's fluency, grammaticality, and coherence.

\section{Results}
\label{sec-results}

\begin{table}
\small
\begin{center}
\begin{tabular}{|l|c|c|c|c|}
\hline
Models & R-1 & R-2 & R-L & M\\
\hline
\multicolumn{5}{|c|}{\textsc{Previous Work}}\\
\hline
Nallapati~\shortcite{nallapati2016abstractive}$^\star$ & 35.46 & 13.30 & 32.65 & - \\
See et al.~\shortcite{see2017get} & 39.53 & 17.28 & 36.38 & 18.72 \\
Paulus~\shortcite{paulus2017deep} {\tiny(XE)}$^\star$ & 38.30 & 14.81 & 35.49 & -  \\
Paulus~\shortcite{paulus2017deep} {\tiny(RL)}$^\star$ & 39.87 & 15.82 & 36.90 & -  \\
\hline
\multicolumn{5}{|c|}{\textsc{Our Models}}\\
\hline
Baseline {\tiny(XE)} & 39.41 & 17.33 & 36.07  & 18.27 \\
ROUGE {\tiny(RL)} & 39.99 & 17.72 & 36.66 & 18.93 \\
Entail {\tiny(RL)} & 39.53 & 17.51 & 36.44 & 20.15  \\
ROUGESal {\tiny(RL)} & 40.36 & 17.97 & 37.00 & 19.84 \\
ROUGE+Ent {\tiny(RL)} & 40.37 & 17.89 & 37.13 & 19.94 \\
ROUGESal+Ent {\tiny(RL)} & 40.43 & 18.00 & 37.10 & 20.02  \\
\hline
\end{tabular}
\end{center}
\vspace{-10pt}
\caption{Results on CNN/Daily Mail (non-anonymous). $^\star$ represents previous work on anonymous version. `XE': cross-entropy loss, `RL': reinforce mixed loss (XE+RL). Columns `R': ROUGE, `M': METEOR.}
\vspace{-5pt}
\label{table:cnndm_non_anonymous_results}
\end{table}

\paragraph{Baseline Cross-Entropy Model Results}
Our abstractive summarization model has attention, pointer-copy, and coverage mechanism. First, we apply cross-entropy optimization and achieve comparable results on CNN/Daily Mail w.r.t. previous work~\cite{see2017get}.\footnote{Our baseline is statistically equal to the paper-reported scores of~\newcite{see2017get} (see Table~\ref{table:cnndm_non_anonymous_results}) on ROUGE-1, ROUGE-2, based on the bootstrap test~\cite{efron1994introduction}. Our baseline is stat. significantly better ($p<0.001$) in all ROUGE metrics w.r.t. the github scores (R-1: 38.82, R-2: 16.81, R-3: 35.71, M: 18.14) of~\newcite{see2017get}.}

\paragraph{ROUGE Reward Results}
First, using ROUGE-L as RL reward (shown as ROUGE in Table~\ref{table:cnndm_non_anonymous_results}) improves the performance on CNN/Daily Mail in all metrics with stat. significant scores ($p<0.001$) as compared to the cross-entropy baseline (and also stat. signif. w.r.t.~\newcite{see2017get}). Similar to~\newcite{paulus2017deep}, we use mixed loss function (XE+RL) for all our reinforcement experiments, to ensure good readability of generated summaries.

\paragraph{ROUGESal and Entail Reward Results}
With our novel ROUGESal reward, we achieve stat. signif. improvements in all metrics w.r.t. the baseline as well as w.r.t. ROUGE-reward results ($p<0.001$), showing that saliency knowledge is strongly improving the summarization model. For our Entail reward, we achieve stat. signif. improvements in ROUGE-L ($p<0.001$) w.r.t. baseline and achieve the best METEOR score by a large margin. See Sec.~\ref{sec:analysis} for analysis of the saliency/entailment skills learned by our models.

\paragraph{Multi-Reward Results}
Similar to ROUGESal, Entail is a better reward when combined with the complementary phrase-matching metric information in ROUGE; Table~\ref{table:cnndm_non_anonymous_results} shows that the ROUGE+Entail multi-reward combination performs stat. signif. better than ROUGE-reward in ROUGE-1, ROUGE-L, and METEOR ($p<0.001$), and better than Entail-reward in all ROUGE metrics. Finally, we combined our two rewards ROUGESal+Entail to incorporate both saliency and entailment knowledge, and it gives the best results overall ($p<0.001$ in all metrics w.r.t. both baseline and ROUGE-reward models), setting the new state-of-the-art.\footnote{Our last three rows in Table~\ref{table:cnndm_non_anonymous_results} are all stat. signif. better in all metrics with $p<0.001$ compared to~\newcite{see2017get}.}

\begin{table}
\small
\begin{center}
\begin{tabular}{|l|c|c|c|c|}
\hline
Models & R-1 & R-2 & R-L & M\\
\hline
Baseline {\tiny(XE)} & 35.50 & 14.57 & 32.19 & 14.36 \\
ROUGE {\tiny(RL)} & 35.97 & 15.45 & 32.72 & 14.50 \\
ROUGESal+Ent {\tiny(RL)} & 38.95 & 17.05 & 35.52 & 16.47 \\
\hline
\end{tabular}
\end{center}
\vspace{-10pt}
\caption{ROUGE F1 full length scores of our models on test-only DUC-2002 generalizability setup.\vspace{-2pt}}
\label{table:duc2002_results}
\end{table}

\begin{table}[t]
\begin{small}
\begin{center}
\begin{tabular}{|l|c|c|c|}
\hline
Models & Relevance & Readability & Total \\
\hline
ROUGESal+Ent & 55 & 54 & 109\\
\newcite{see2017get} & 34 & 33 & 67 \\
Non-distinguish. & 11 & 13 & 24 \\
\hline
\end{tabular}
\end{center}
\vspace{-10pt}
\caption{Human Evaluation: pairwise comparison of relevance and readability between our ROUGESal+Entail multi-reward model and ~\newcite{see2017get}.
}
\label{table:human-eval-results}
\vspace{-6pt}
\end{small}
\end{table}

\paragraph{Human Evaluation}
Table.~\ref{table:human-eval-results} shows the MTurk anonymous human evaluation study (based on $100$ samples), where we do pairwise comparison between our ROUGESal+Entail multi-reward's output summaries w.r.t.~\newcite{see2017get} summaries on  CNN/Daily Mail (see setup details in Sec.~\ref{subsec:evaluation-metrics}). As shown, our multi-reward model is better on both relevance and readability.

\paragraph{Test-Only Transfer (DUC-2002) Results} 
Finally, we also tested our model's generalizability/transfer skills, where we take the models trained on CNN/Daily Mail and directly test them on DUC-2002 in a test-only setup. As shown in Table~\ref{table:duc2002_results}, our final ROUGESal+Entail multi-reward RL model is statistically significantly better than both the cross-entropy (pointer-generator + coverage) baseline as well as ROUGE reward RL model, in terms of all 4 metrics with a large margin (with $p<0.001$). This demonstrates that our ROUGESal+Entail model learned better transferable and generalizable skills of saliency and logical entailment.

\section{Output Analysis}
\label{sec:analysis}

\paragraph{Saliency Analysis} We analyzed the output summaries generated by~\newcite{see2017get}, and our baseline, ROUGE-reward and ROUGESal-reward models, using our saliency prediction model (Sec.~\ref{subsec:saliency-rewards}) as the keyword detection classifier. We annotated the ground-truth and model summaries with this keyword classifier and computed the \% match, i.e., how many salient words from the ground-truth summary were also generated in the model summary\footnote{In order to select the keywords for this analysis, we used a $0.2$ probability threshold on the saliency classifier (based on the scale of the classifier's distribution).}, and the scores are $27.95\%$, $28.00\%$, $28.80\%$, and $30.86\%$. We also used the original CNN/Daily Mail Cloze Q\&A setup~\cite{hermann2015teaching} with the fill-in-the-blank answers treated as salient information, and the results are $60.66\%$, $59.36\%$, $60.67\%$, and $64.66\%$ for the four models. Further, we also calculated the ROUGESal scores (based on our reward formulation in Sec.~\ref{subsec:saliency-rewards}), and the results are $42.04\%$, $42.14\%$, $43.05\%$, and $46.56\%$ for the four models. All three of these saliency analysis experiments illustrate that our ROUGESal reward model is stat. signif. better in saliency than the~\newcite{see2017get}, our baseline, and ROUGE-reward models ($p<0.001$ for all three experiments).

\paragraph{Entailment Analysis}
We also analyzed the entailment scores of the generated summaries from~\newcite{see2017get}, and our baseline, ROUGE-reward, and Entail-reward models, and the results are $27.33\%$, $27.21\%$, $28.23\%$, and $28.98\%$.\footnote{Based on our ground-truth summary to output summary sentences' average entailment score (see Sec.~\ref{subsec:entailment-rewards}); similar trends hold for document-to-summary entailment scores.} 
We observe that our Entail-reward model achieves stat. significant entailment scores ($p<0.001$) w.r.t. all the other three models.

\begin{table}[t]
\small
\begin{center}
\begin{tabular}{|l|c|c|c|}
\hline
Models  & 2-gram & 3-gram & 4-gram \\
\hline
\newcite{see2017get} & 2.24 & 6.03 & 9.72 \\
Baseline {\tiny(XE)} &  2.23 & 5.58 & 8.81 \\
ROUGE {\tiny(RL)} &  2.69 & 6.57 & 10.23 \\
ROUGESal {\tiny(RL)}  & 2.37 & 6.00 & 9.50 \\
Entail {\tiny(RL)} & 2.63  & 6.56 & 10.26 \\
\hline
\end{tabular}
\end{center}
\vspace{-10pt}
\caption{Abstractiveness: novel $n$-gram percentage.}
\label{table:novel-n-gram}
\vspace{-10pt}
\end{table}

\paragraph{Abstractiveness Analysis} In order to measure the abstractiveness of our models, we followed the `novel $n$-gram counts' approach suggested in~\newcite{see2017get}. First, we found that all our reward-based RL models have significantly ($p<0.01$) more novel $n$-grams than our cross-entropy baseline (see Table~\ref{table:novel-n-gram}). Next, the Entail-reward model `maintains' stat. equal abstractiveness as the ROUGE-reward model, likely because it encourages rewriting to create logical subsets of information, while the ROUGESal-reward model does a bit worse, probably because it focuses on copying more salient information (e.g., names). Compared to previous work~\cite{see2017get}, our Entail-reward and ROUGE-reward models achieve statistically significant improvement ($p<0.01$) while ROUGESal is comparable.

\section{Conclusion}
We presented a summarization model trained with novel RL reward functions to improve the saliency and directed logical entailment aspects of a good summary. Further, we introduced the novel and effective multi-reward approach of optimizing multiple rewards simultaneously in alternate mini-batches. We achieve the new state-of-the-art on CNN/Daily Mail and also strong test-only improvements on a DUC-2002 transfer setup.

\section*{Acknowledgments}
We thank the reviewers for their helpful comments. This work was supported by DARPA (YFA17-D17AP00022), Google Faculty Research Award, Bloomberg Data Science Research Grant, and NVidia GPU awards. The views, opinions, and/or findings contained in this article are those of the authors and should not be interpreted as representing the official views or policies, either expressed or implied, of the funding agency.

\bibliography{citations}

\begin{thebibliography}{}
\expandafter\ifx\csname natexlab\endcsname\relax\def\natexlab#1{#1}\fi

\bibitem[{Bahdanau et~al.(2015)Bahdanau, Cho, and Bengio}]{bahdanau2014neural}
Dzmitry Bahdanau, Kyunghyun Cho, and Yoshua Bengio. 2015.
\newblock Neural machine translation by jointly learning to align and
  translate.
\newblock In {\em ICLR\/}.

\bibitem[{Bowman et~al.(2015)Bowman, Angeli, Potts, and
  Manning}]{bowman2015large}
Samuel~R Bowman, Gabor Angeli, Christopher Potts, and Christopher~D Manning.
  2015.
\newblock A large annotated corpus for learning natural language inference.
\newblock In {\em EMNLP\/}.

\bibitem[{Chen et~al.(2016)Chen, Zhu, Ling, Wei, and
  Jiang}]{Chen2016DistractionBasedNN}
Qian Chen, Xiaodan Zhu, Zhenhua Ling, Si~Wei, and Hui Jiang. 2016.
\newblock Distraction-based neural networks for modeling documents.
\newblock In {\em IJCAI\/}.

\bibitem[{Cheung and Penn(2014)}]{cheung2014unsupervised}
Jackie Chi~Kit Cheung and Gerald Penn. 2014.
\newblock Unsupervised sentence enhancement for automatic summarization.
\newblock In {\em EMNLP\/}. pages 775--786.

\bibitem[{Chopra et~al.(2016)Chopra, Auli, and Rush}]{chopra2016abstractive}
Sumit Chopra, Michael Auli, and Alexander~M Rush. 2016.
\newblock Abstractive sentence summarization with attentive recurrent neural
  networks.
\newblock In {\em HLT-NAACL\/}.

\bibitem[{Clarke and Lapata(2008)}]{clarke2008global}
James Clarke and Mirella Lapata. 2008.
\newblock Global inference for sentence compression: An integer linear
  programming approach.
\newblock {\em Journal of Artificial Intelligence Research\/} 31:399--429.

\bibitem[{Dagan et~al.(2006)Dagan, Glickman, and Magnini}]{dagan2006pascal}
Ido Dagan, Oren Glickman, and Bernardo Magnini. 2006.
\newblock The pascal recognising textual entailment challenge.
\newblock In {\em Machine learning challenges. evaluating predictive
  uncertainty, visual object classification, and recognising tectual
  entailment\/}, Springer, pages 177--190.

\bibitem[{Denkowski and Lavie(2014)}]{banerjee2005meteor}
Michael Denkowski and Alon Lavie. 2014.
\newblock Meteor universal: Language specific translation evaluation for any
  target language.
\newblock In {\em EACL\/}.

\bibitem[{Dohare and Karnick(2017)}]{dohare2017text}
Shibhansh Dohare and Harish Karnick. 2017.
\newblock Text summarization using abstract meaning representation.
\newblock {\em arXiv preprint arXiv:1706.01678\/} .

\bibitem[{Efron and Tibshirani(1994)}]{efron1994introduction}
Bradley Efron and Robert~J Tibshirani. 1994.
\newblock {\em An introduction to the bootstrap\/}.
\newblock CRC press.

\bibitem[{Filippova et~al.(2015)Filippova, Alfonseca, Colmenares, Kaiser, and
  Vinyals}]{filippova2015sentence}
Katja Filippova, Enrique Alfonseca, Carlos~A Colmenares, Lukasz Kaiser, and
  Oriol Vinyals. 2015.
\newblock Sentence compression by deletion with lstms.
\newblock In {\em EMNLP\/}. pages 360--368.

\bibitem[{Ganesan et~al.(2010)Ganesan, Zhai, and Han}]{ganesan2010opinosis}
Kavita Ganesan, ChengXiang Zhai, and Jiawei Han. 2010.
\newblock Opinosis: a graph-based approach to abstractive summarization of
  highly redundant opinions.
\newblock In {\em Proceedings of the 23rd international conference on
  computational linguistics\/}. ACL, pages 340--348.

\bibitem[{Gerani et~al.(2014)Gerani, Mehdad, Carenini, Ng, and
  Nejat}]{gerani2014abstractive}
Shima Gerani, Yashar Mehdad, Giuseppe Carenini, Raymond~T Ng, and Bita Nejat.
  2014.
\newblock Abstractive summarization of product reviews using discourse
  structure.
\newblock In {\em EMNLP\/}. volume~14, pages 1602--1613.

\bibitem[{Giannakopoulos(2009)}]{giannakopoulos2009automatic}
George Giannakopoulos. 2009.
\newblock Automatic summarization from multiple documents.
\newblock {\em Ph. D. dissertation\/} .

\bibitem[{Gupta et~al.(2014)Gupta, Kaur, Singh, Goel, and
  Mirkin}]{gupta2014text}
Anand Gupta, Manpreet Kaur, Adarsh Singh, Aseem Goel, and Shachar Mirkin. 2014.
\newblock Text summarization through entailment-based minimum vertex cover.
\newblock {\em Lexical and Computational Semantics (* SEM 2014)\/} page~75.

\bibitem[{Harabagiu and Hickl(2006)}]{harabagiu2006methods}
Sanda Harabagiu and Andrew Hickl. 2006.
\newblock Methods for using textual entailment in open-domain question
  answering.
\newblock In {\em ACL\/}. pages 905--912.

\bibitem[{Hermann et~al.(2015)Hermann, Kocisky, Grefenstette, Espeholt, Kay,
  Suleyman, and Blunsom}]{hermann2015teaching}
Karl~Moritz Hermann, Tomas Kocisky, Edward Grefenstette, Lasse Espeholt, Will
  Kay, Mustafa Suleyman, and Phil Blunsom. 2015.
\newblock Teaching machines to read and comprehend.
\newblock In {\em NIPS\/}. pages 1693--1701.

\bibitem[{Isonuma et~al.(2017)Isonuma, Fujino, Mori, Matsuo, and
  Sakata}]{isonuma2017extractive}
Masaru Isonuma, Toru Fujino, Junichiro Mori, Yutaka Matsuo, and Ichiro Sakata.
  2017.
\newblock Extractive summarization using multi-task learning with document
  classification.
\newblock In {\em EMNLP\/}. pages 2091--2100.

\bibitem[{Jimenez et~al.(2014)Jimenez, Duenas, Baquero, Gelbukh, B{\'a}tiz, and
  Mendiz{\'a}bal}]{jimenez2014unal}
Sergio Jimenez, George Duenas, Julia Baquero, Alexander Gelbukh, Av~Juan~Dios
  B{\'a}tiz, and Av~Mendiz{\'a}bal. 2014.
\newblock {UNAL-NLP}: Combining soft cardinality features for semantic textual
  similarity, relatedness and entailment.
\newblock In {\em In {SemEval}\/}. pages 732--742.

\bibitem[{Jing(2000)}]{jing2000sentence}
Hongyan Jing. 2000.
\newblock Sentence reduction for automatic text summarization.
\newblock In {\em ANLP\/}.

\bibitem[{Kingma and Ba(2015)}]{kingma2014adam}
Diederik Kingma and Jimmy Ba. 2015.
\newblock Adam: A method for stochastic optimization.
\newblock In {\em ICLR\/}.

\bibitem[{Knight and Marcu(2002)}]{knight2002summarization}
Kevin Knight and Daniel Marcu. 2002.
\newblock Summarization beyond sentence extraction: A probabilistic approach to
  sentence compression.
\newblock {\em Artificial Intelligence\/} 139(1):91--107.

\bibitem[{Lai and Hockenmaier(2014)}]{lai2014illinois}
Alice Lai and Julia Hockenmaier. 2014.
\newblock Illinois-{LH}: A denotational and distributional approach to
  semantics.
\newblock {\em Proc. SemEval\/} 2:5.

\bibitem[{Lin(2004)}]{lin2004rouge}
Chin-Yew Lin. 2004.
\newblock {ROUGE}: A package for automatic evaluation of summaries.
\newblock In {\em Text Summarization Branches Out: Proceedings of the ACL-04
  workshop\/}. volume~8.

\bibitem[{Liu et~al.(2015)Liu, Flanigan, Thomson, Sadeh, and
  Smith}]{liu2015toward}
Fei Liu, Jeffrey Flanigan, Sam Thomson, Norman Sadeh, and Noah~A Smith. 2015.
\newblock Toward abstractive summarization using semantic representations.
\newblock In {\em NAACL: HLT\/}. pages 1077--1086.

\bibitem[{Mehdad et~al.(2013)Mehdad, Carenini, Tompa, and
  Ng}]{mehdad2013abstractive}
Yashar Mehdad, Giuseppe Carenini, Frank~W Tompa, and Raymond~T Ng. 2013.
\newblock Abstractive meeting summarization with entailment and fusion.
\newblock In {\em Proc. of the 14th European Workshop on Natural Language
  Generation\/}. pages 136--146.

\bibitem[{Nallapati et~al.(2016)Nallapati, Zhou, Gulcehre, Xiang
  et~al.}]{nallapati2016abstractive}
Ramesh Nallapati, Bowen Zhou, Caglar Gulcehre, Bing Xiang, et~al. 2016.
\newblock Abstractive text summarization using sequence-to-sequence rnns and
  beyond.
\newblock In {\em CoNLL\/}.

\bibitem[{Parikh et~al.(2016)Parikh, T{\"a}ckstr{\"o}m, Das, and
  Uszkoreit}]{parikh2016decomposable}
Ankur~P Parikh, Oscar T{\"a}ckstr{\"o}m, Dipanjan Das, and Jakob Uszkoreit.
  2016.
\newblock A decomposable attention model for natural language inference.
\newblock In {\em EMNLP\/}.

\bibitem[{Pasunuru and Bansal(2017)}]{pasunuru2017reinforced}
Ramakanth Pasunuru and Mohit Bansal. 2017.
\newblock Reinforced video captioning with entailment rewards.
\newblock In {\em EMNLP\/}.

\bibitem[{Paulus et~al.(2017)Paulus, Xiong, and Socher}]{paulus2017deep}
Romain Paulus, Caiming Xiong, and Richard Socher. 2017.
\newblock A deep reinforced model for abstractive summarization.
\newblock {\em arXiv preprint arXiv:1705.04304\/} .

\bibitem[{Rajpurkar et~al.(2016)Rajpurkar, Zhang, Lopyrev, and
  Liang}]{rajpurkar2016squad}
Pranav Rajpurkar, Jian Zhang, Konstantin Lopyrev, and Percy Liang. 2016.
\newblock Squad: 100,000+ questions for machine comprehension of text.
\newblock In {\em EMNLP\/}.

\bibitem[{Ranzato et~al.(2015)Ranzato, Chopra, Auli, and
  Zaremba}]{ranzato2015sequence}
Marc'Aurelio Ranzato, Sumit Chopra, Michael Auli, and Wojciech Zaremba. 2015.
\newblock Sequence level training with recurrent neural networks.
\newblock In {\em ICLR\/}.

\bibitem[{Rennie et~al.(2016)Rennie, Marcheret, Mroueh, Ross, and
  Goel}]{rennie2016self}
Steven~J Rennie, Etienne Marcheret, Youssef Mroueh, Jarret Ross, and Vaibhava
  Goel. 2016.
\newblock Self-critical sequence training for image captioning.
\newblock {\em arXiv preprint arXiv:1612.00563\/} .

\bibitem[{Rush et~al.(2015)Rush, Chopra, and Weston}]{rush2015neural}
Alexander~M Rush, Sumit Chopra, and Jason Weston. 2015.
\newblock A neural attention model for abstractive sentence summarization.
\newblock In {\em CoRR\/}.

\bibitem[{See et~al.(2017)See, Liu, and Manning}]{see2017get}
Abigail See, Peter~J Liu, and Christopher~D Manning. 2017.
\newblock Get to the point: Summarization with pointer-generator networks.
\newblock In {\em ACL\/}.

\bibitem[{Subramanian et~al.(2017)Subramanian, Wang, Yuan, and
  Trischler}]{subramanian2017neural}
Sandeep Subramanian, Tong Wang, Xingdi Yuan, and Adam Trischler. 2017.
\newblock Neural models for key phrase detection and question generation.
\newblock {\em arXiv preprint arXiv:1706.04560\/} .

\bibitem[{Suzuki and Nagata(2016)}]{Suzuki2016Summ}
Jun Suzuki and Masaaki Nagata. 2016.
\newblock Rnn-based encoder-decoder approach with word frequency estimation.
\newblock In {\em EACL\/}.

\bibitem[{Wang et~al.(2016)Wang, Raghavan, Castelli, Florian, and
  Cardie}]{wang2016sentence}
Lu~Wang, Hema Raghavan, Vittorio Castelli, Radu Florian, and Claire Cardie.
  2016.
\newblock A sentence compression based framework to query-focused
  multi-document summarization.
\newblock {\em arXiv preprint arXiv:1606.07548\/} .

\bibitem[{Williams et~al.(2017)Williams, Nangia, and
  Bowman}]{williams2017broad}
Adina Williams, Nikita Nangia, and Samuel~R Bowman. 2017.
\newblock A broad-coverage challenge corpus for sentence understanding through
  inference.
\newblock {\em arXiv preprint arXiv:1704.05426\/} .

\bibitem[{Williams(1992)}]{williams1992simple}
Ronald~J Williams. 1992.
\newblock Simple statistical gradient-following algorithms for connectionist
  reinforcement learning.
\newblock {\em Machine learning\/} 8(3-4):229--256.

\bibitem[{Wu et~al.(2016)Wu, Schuster, Chen, Le, Norouzi, Macherey, Krikun,
  Cao, Gao, Macherey et~al.}]{wu2016google}
Yonghui Wu, Mike Schuster, Zhifeng Chen, Quoc~V Le, Mohammad Norouzi, Wolfgang
  Macherey, Maxim Krikun, Yuan Cao, Qin Gao, Klaus Macherey, et~al. 2016.
\newblock Google's neural machine translation system: Bridging the gap between
  human and machine translation.
\newblock {\em arXiv preprint arXiv:1609.08144\/} .

\bibitem[{Zaremba and Sutskever(2015)}]{zaremba2015reinforcement}
Wojciech Zaremba and Ilya Sutskever. 2015.
\newblock Reinforcement learning neural turing machines.
\newblock {\em arXiv preprint arXiv:1505.00521\/} 362.

\bibitem[{Zhang et~al.(2004)Zhang, Zincir-Heywood, and Milios}]{zhang2004world}
Yongzheng Zhang, Nur Zincir-Heywood, and Evangelos Milios. 2004.
\newblock World wide web site summarization.
\newblock {\em Web Intelligence and Agent Systems: An International Journal\/}
  2(1):39--53.

\end{thebibliography}
\bibliographystyle{acl_natbib}

\appendix
\section{Supplementary Material}
\subsection{Saliency Rewards}
\label{suppl:subsec:saliency-rewards}
Here, we describe the ROUGE-L formulation at summary-level and later describe how we incorporate saliency information into it. Given a reference summary of $u$ sentences containing a total of $m$ tokens ($\{w_{r,k}\}_{k=1}^m$) and a generated summary of $v$ sentences with a total of $n$ tokens ($\{w_{c,k}\}_{k=1}^n$), let $r_i$ be the reference summary sentence and $c_j$ be the generated summary sentence. Then, the precision ($P_{lcs}$), recall ($R_{lcs}$), and F-score ($F_{lcs}$) for ROUGE-L are defined as follows:
\begin{equation}
P_{lcs} = \frac{\sum_{i=1}^u LCS_{\cup} (r_i,C)}{n}
\end{equation}
\begin{equation}
R_{lcs} = \frac{\sum_{i=1}^u LCS_{\cup} (r_i,C)}{m}
\end{equation}
\begin{equation}
F_{lcs} = \frac{(1+\beta^2)R_{lcs}P_{lcs}}{R_{lcs}+\beta^2 P_{lcs}}
\end{equation}
where $LCS_{\cup}$ takes the \emph{union} Longest Common Subsequence (LCS) between a reference summary sentence $r_i$ and every generated summary sentence $c_j$ ($c_j \in C$), and $\beta$ is defined in~\newcite{lin2004rouge}. In the above ROUGE-L scores, we assume that every token has equal weight, i.e, $1$. However, every summary has salient tokens which should be rewarded with more weight. Hence, we use the weights obtained from our novel saliency predictor to modify the ROUGE-L scores with salient information as follows:
\begin{equation}
P^s_{lcs} = \frac{\sum_{i=1}^u LCS^*_{\cup} (r_i,C)}{\sum_{k=1}^n \eta(w_{c,k})}
\end{equation}
\begin{equation}
R^s_{lcs} = \frac{\sum_{i=1}^u LCS^*_{\cup} (r_i,C)}{\sum_{k=1}^m \eta(w_{r,k})}
\end{equation}
\begin{equation}
F^s_{lcs} = \frac{(1+\beta^2)R^s_{lcs}P^s_{lcs}}{R^s_{lcs}+\beta^2 P^s_{lcs}}
\end{equation}
where $\eta(w)$ is the weight assigned by the saliency predictor for token $w$, and $\beta$ is defined in~\newcite{lin2004rouge}.\footnote{If a token is repeated at multiple times in the input sentence, we average the probabilities of those instances.} Let $\{w_k\}_{k=1}^p$ be the union LCS set, then $LCS^*_{\cup}(r_i,C)$ is defined as follows:
\begin{equation}
LCS^*_{\cup}(r_i,C) = \sum_{k=1}^p \eta(w_k) 
\end{equation}

\subsection{Experimental Setup}
\label{suppl:sec-setup}

\subsubsection{Datasets}
\label{subsec:suppl:datasets}

\paragraph{CNN/Daily Mail Dataset}
CNN/Daily Mail dataset~\cite{hermann2015teaching,nallapati2016abstractive} is a collection of online articles and their summaries. The summaries are based on the human written highlights of these articles. The dataset has $287,226$ training pairs, $13,368$ validation pairs, and $11,490$ test pairs. We use the non-anonymous version of the dataset as described in~\newcite{see2017get}.
 
\paragraph{DUC Test Corpus}
We use the DUC-2002 single document summarization dataset\footnote{\scriptsize{\url{http://www-nlpir.nist.gov/projects/duc/guidelines/2002.html}}} as a test-only setup where we directly take the pretrained models trained on CNN/Daily Mail dataset and test them on DUC-2002, in order to check for our model's domain transfer capabilities. This corpus consists of $567$ documents with one or two human annotated reference summaries.

\paragraph{SNLI and MultiNLI corpus}
We use the full Stanford Natural Language Inference (SNLI) corpus~\cite{bowman2015large} and the recent Multi-NLI corpus~\cite{williams2017broad} data for building our entailment classifier. We use the standard splits following previous work.

\paragraph{SQuAD Dataset}
We use Stanford Question Answering Dataset (SQuAD) for our saliency prediction model. We process the SQuAD dataset to collect the sentence and their corresponding salient phrases pairs. Here again, we use the standard split following previous work.

\subsubsection{Training Details}
\label{subsec:suppl:training-details}
During training, all our LSTM-RNNs are set with hidden state size of $256$. We use a vocabulary size of $50\textrm{k}$, where word embeddings are represented in $128$ dimension, and both the encoder and decoder share the same embedding for each word. We encode the source document using a $400$ time-step unrolled LSTM-RNN and $100$ time-step unrolled LSTM-RNN for decoder. We clip the gradients to a maximum gradient norm value of $2.0$ and use Adam optimizer~\cite{kingma2014adam} with a learning rate of $1 \times 10^{-3}$ for pointer baseline and $1 \times 10^{-4}$ while training along with coverage loss, and $1 \times 10^{-6}$ for reinforcement learning. Following~\newcite{see2017get}, we add coverage mechanism to a converged pointer model. For mixed-loss (XE+RL) optimization, we use the following $\gamma$ values for various rewards: $0.9985$ for ROUGE, $0.9999$ for Entail and ROUGE+Entail, and $0.9995$ for ROUGESal and ROUGESal+Entail. For reinforcement learning, we only use $5000$ training samples ($< 2\%$ of the actual data) to speed up convergence, but we found it to work well in practice. During inference time, we use a beam search of size $4$.

\subsection{Results}

\subsubsection{Saliency and Entailment Scorer}
Table~\ref{table:entailment_saliency_predictor} presents the performance of our saliency predictor (on the SQuAD-based dev set for answer span classification accuracy) and entailment classifier (on the Multi-NLI dev set accuracy). Our entailment classifier is comparable to the state-of-the-art models.\footnote{RepEval leaderboard: \url{https://repeval2017.github.io/shared/}} 
\begin{table}
\begin{center}
\begin{tabular}{|l|c|}
\hline
Models & Accuracy \\
\hline
Entailment Classifier & 74.50\% \\
Saliency Predictor & 16.87\% \\
\hline
\end{tabular}
\end{center}
\caption{Performance of our entailment classifier and saliency predictor.}
\label{table:entailment_saliency_predictor}
\end{table}

\end{document}